\documentclass[conference]{IEEEtran}

\usepackage{cite}
\usepackage{amsmath,amssymb,amsfonts}
\usepackage{graphicx}
\usepackage{booktabs}
\usepackage{array}
\usepackage[utf8]{inputenc}
\usepackage[T1]{fontenc}
\usepackage{hyperref}
\hypersetup{colorlinks=true,linkcolor=blue,citecolor=blue,urlcolor=blue}

\title{Skill-Contracted Agents for Evidence-Aware Materials Literature Analysis}

\author{
\IEEEauthorblockN{Bixuan Li$^{1,2}$, Yu Liu$^{*,1,2,3}$, Shuo Shi$^{1}$, Xiaoya Huang$^{1,2}$, Peng Kang$^{*,1,2,3}$, and Lei Zheng$^{*,1,2,3}$}
\IEEEauthorblockA{$^{1}$Tianmushan Laboratory, Hangzhou 311115, P. R. China.}
\IEEEauthorblockA{$^{2}$National Key Laboratory of Artificial Intelligence for Materials Science, Beihang University, Beijing 100191, P. R. China.}
\IEEEauthorblockA{$^{3}$School of Materials Science and Engineering, Beihang University, Beijing 100191, P. R. China.}
\IEEEauthorblockA{$^{*}$Corresponding authors: zhenglei@buaa.edu.cn, liuyucarbon@buaa.edu.cn, pengkang@buaa.edu.cn}
}

\begin{document}
\maketitle

\begin{abstract}
Materials science literature analysis requires simultaneous attention to composition, processing, characterization, and property relationships, yet conventional retrieval-augmented generation pipelines struggle to reconcile heterogeneous tasks within a single retrieve-then-generate architecture. Here we present AlphaAgent, a skill-driven agent framework that decouples retrieval-based question answering from paper-level report generation through explicit skill contracts. A dedicated retrieval skill rewrites user requests into material-specific search intents, queries a curated index of more than 300,000 papers from the Journal Citation Reports Metallurgy and Metallurgical Engineering category, and reformulates queries when initial evidence is insufficient. A separate report-generation skill parses full-text PDFs to produce structured per-paper analytical reports and cross-paper summaries. In a blind evaluation on 40 materials-science questions, half of which required deep analytical reasoning, AlphaAgent substantially outperformed a baseline system matched for underlying model, document index, and retrieval scale, with the largest gains in mechanistic explanation and awareness of credibility boundaries. These results indicate that explicit task separation, refined retrieval intent, and evidence-aware generation improve large-language-model-based literature analysis for materials research.
\end{abstract}

\section{Introduction}

Progress in materials science depends on the cumulative interpretation of published evidence. Many questions in materials science are not resolved by a single study; they require comparison across composition, processing history, characterization, and testing conditions, all of which can change how a reported mechanism or property should be interpreted~\cite{reed2006superalloys,miracle2017critical}. The accelerating growth of the materials literature has turned such evidence synthesis into a practical bottleneck~\cite{bornmann2021growth}, especially for questions that require both factual retrieval and mechanistic comparison across multiple studies. Recent advances in large language models (LLMs) have created new opportunities for materials literature analysis~\cite{tshitoyan2019unsupervised,dagdelen2024structured,choi2024accelerating,lei2024materials}. Unlike conventional retrieval or task-specific text-mining pipelines, LLMs can read unstructured scientific text in a question-conditioned manner, relate evidence expressed across heterogeneous terminology and reporting styles, and organize relevant findings into coherent responses for downstream interpretation. When coupled with retrieval over external sources, they provide a practical basis for literature-analysis systems that can operate at the scale and complexity of modern materials science. 

Current LLM-based approaches still face important limitations in scientific literature analysis. Without external grounding, general-purpose LLMs may produce factually unsupported statements, a limitation that is especially problematic in materials science, where interpretation depends strongly on experimental context~\cite{ji2023survey}. Web-browsing and search-grounded systems reduce this risk by linking answers to online sources~\cite{nakano2021webgpt,vu2024freshllms}, but open web search does not distinguish consistently between peer-reviewed primary studies, unreviewed preprints, and secondary or popularized summaries. Retrieval-augmented generation (RAG) over a curated document index addresses part of this source-control problem~\cite{lewis2020retrieval,huang2025survey}, but important difficulties remain in materials-science literature analysis. Materials terminology is strongly context dependent: the same phase name, processing term, or property descriptor can carry different meanings across material systems and characterization settings, introducing ambiguity into semantic retrieval~\cite{weston2019named}. A further difficulty arises at the ranking stage, where matched passages are often drawn disproportionately from introductions and abstracts, leading LLMs to ground their answers in broad and shallow summaries rather than in more specific analytical evidence~\cite{liu2024lost}. Beyond retrieval, current systems also remain limited in the form of scientific synthesis they provide. Most RAG pipelines or general LLMs return short, answer-style summaries, but rigorous literature work in materials science often requires structured, figure-aware reading of full papers, including extraction of experimental protocols, microstructural evidence, and mechanistic interpretation from the main text, figures, and captions~\cite{xu2020layoutlm,appalaraju2021docformer}.

Here we introduce AlphaAgent, a skill-driven framework for materials-science literature analysis that couples retrieval-grounded question answering with structured paper-level reading. AlphaAgent decomposes literature work into two contract-bound skills: a retrieval skill that produces evidence-controlled answers from a curated literature index, and a report-generation skill that turns validated paper sets into document-level analytical reports~\cite{yao2023react,schick2023toolformer,li2023camel,buehler2024generative}. The retrieval skill converts user questions into materials-specific search intents that keep the target material system, experimental context, property of interest and analytical focus linked during retrieval. It then evaluates whether the returned evidence is sufficiently aligned with this intent, and reformulates the query based on both the original intent and the observed evidence gap when the evidence is too broad, shallow or scientifically mismatched. The report-generation skill, in turn, extends the validated paper set into structured reading outputs by parsing full-text PDFs and generating per-paper analytical reports and cross-paper syntheses. These reports organize reported findings, evidentiary scope and reading paths to support evidence-grounded interpretation beyond short summary-style output. We evaluate AlphaAgent's generated answers on 40 materials-science questions spanning general conceptual and deep analytical tasks. Under matched conditions---the same model, document index and retrieval scale---AlphaAgent substantially improves over a conventional RAG baseline on deep analytical questions, with the largest gains in mechanistic explanation and credibility-boundary awareness. Together, these results show that explicit task separation, retrieval-intent refinement and evidence-aware answer generation can make LLM-based literature analysis more reliable for materials research.

\section{Framework Overview and Skill-Driven Design}

\subsection{System Organization}

AlphaAgent is a task executor for materials science literature analysis. The runtime interprets user requests and routes them to skills whose contracts define expected inputs, output artifacts, permissible tools, and failure behavior. The framework provides two core capabilities: retrieval-based question answering converts user questions into search-oriented queries, retrieves evidence from a materials literature index, and grounds answers in snippets and paper metadata; paper-report generation analyzes selected papers at the document level to produce single-paper reports and cross-paper summaries. The two skills compose into a pipeline in which retrieval precedes report generation, and the resulting query record supplies the paper references and document identifiers used by the report stage. This design builds on recent work on tool-using language models and agentic workflows~\cite{yao2023react,schick2023toolformer,wu2023autogen}, with skill boundaries shaped around the practical structure of literature work.

A skill is a bounded agent capability defined by a semantic input contract, an output artifact schema, permitted evidence sources and tools, validation conditions, and failure behavior. A workflow orders operations; a skill contract constrains what evidence may be used, what intermediate state is valid, and how downstream stages consume the produced artifacts. This distinction matters for scientific literature analysis because it preserves the evidential basis of conclusions across retrieval, document reading, and synthesis stages. By binding each stage to explicit evidence constraints, the framework maintains the integrity of the interpretive chain from source material to final output.

\subsection{Skill Contracts}

Each skill operates under a contract that specifies what it may read, what artifacts it must write, and which operations remain outside its scope. The retrieval skill produces a query result containing the raw user question, rewritten retrieval query, snippets, paper metadata, document references, and citation information. When multiple retrieval attempts are made, the agent selects the best-aligned evidence as the \textit{promoted query result}. This artifact contains the original question, the rewritten retrieval intent, the selected snippets and paper records, document references, citation metadata, and the evidence-assessment outcome. Only this promoted artifact supports answer generation or supplies paper references for report generation.

The report-generation skill accepts a promoted query result, paper references, and document identifiers, then produces a batch report record, per-paper deep-report drafts, rendered reports, and a query-level summary when enough individual reports succeed. This contract-based design separates semantic decisions from runtime mechanics. The agent interprets user requests, writes retrieval intents, and organizes report content; the runtime materializes query attempts, resolves PDF locations, prepares document-analysis workspaces, validates drafts, and renders HTML outputs. This separation ensures that reports are generated only from validated intermediate states, preserving a clear audit trail from user question to final output.

\subsection{Workflow Composition}

\begin{figure*}[t]
\centering
\includegraphics[width=0.98\textwidth]{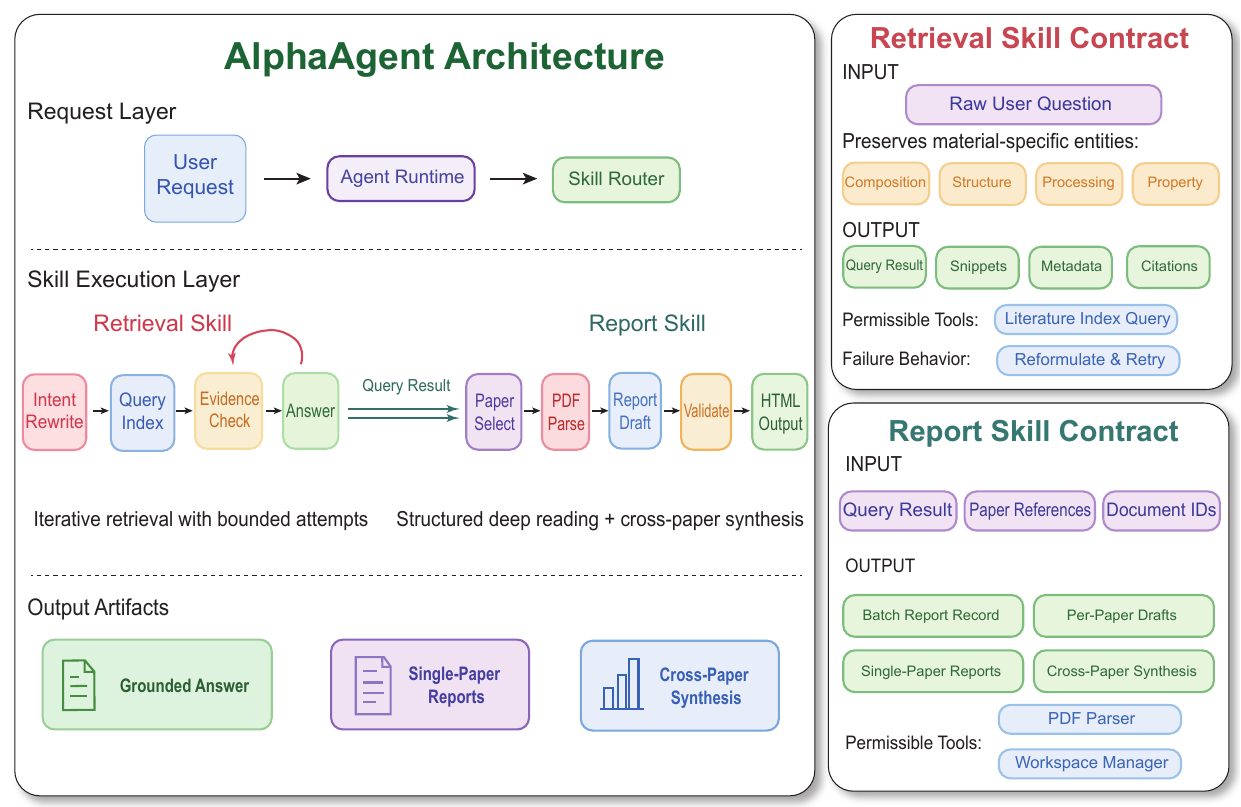}
\caption{AlphaAgent architecture for skill-driven materials literature analysis. The agent runtime routes user requests from the request layer to bounded skills through explicit contracts. Retrieval-based question answering and report generation remain separate execution paths, but they can be composed by passing a query result, paper references, and document identifiers from retrieval to document-level report generation.}
\label{fig:architecture}
\end{figure*}

AlphaAgent routes direct questions to the retrieval skill and returns a source-grounded answer. For requests that explicitly ask for deep reports, the system executes retrieval first and then passes the query result, paper references, and document identifiers to the report-generation skill. The report skill reuses the existing query result for paper selection, maintaining a clear link between the user's original question, retrieved evidence, and subsequent document-level analysis. This design enforces clear boundaries: if retrieval does not yield valid evidence, report generation does not proceed, and individual papers that cannot be parsed do not invalidate the remaining reports. Skill composition achieves both modularity and consistent output behavior, as shown in Figure~\ref{fig:architecture}.

\section{Intent-Driven Iterative Retrieval}

\subsection{Retrieval Intent Construction}

The retrieval skill rewrites user requests into English retrieval intents, following the broader motivation of query rewriting for retrieval-augmented LLMs~\cite{ma2023query}. It strips conversational phrasing, compresses compound questions into search-oriented expressions, and preserves material-specific entities, including alloy grades, phase names, processing parameters, temperature ranges, property terms, and standards. The rewritten intent is stored alongside the raw question so that answer generation remains aligned with the original request. This dual-storage design prevents semantic drift during multi-turn retrieval and ensures that the final answer can be traced back to the user's initial objective. By separating the search-oriented expression from the task-level reference, the framework maintains both retrieval precision and response fidelity across multiple query iterations. Without this separation, dense retrieval favors overly broad terminology, while answer generation loses sight of the specific material system or property trade-off that motivated the query.

Small lexical changes in materials science can alter the retrieved literature scope substantially, because the same phase name or property descriptor may carry different meanings across alloy systems and characterization settings. When a question is converted into a retrieval intent, it must retain its deformation mechanisms, temperature regimes, and microstructural evolution pathways. A query about creep resistance of superalloys, for example, must keep dislocation climb, diffusion, and $\gamma'$ rafting explicit. The rewritten query must also remain concise so that dense retrieval remains effective. AlphaAgent therefore uses the rewritten intent as search text while keeping the raw question as the task-level reference, balancing the need for material-specific detail against the constraints of dense vector search. An overly verbose intent dilutes the signal in dense vector search; an overly abstract intent yields passages that lack the mechanistic detail required for materials-specific answers.

\subsection{Iterative Retrieval and Attempt Selection}

Conventional RAG typically performs a single retrieval pass before generating an answer from returned passages, which limits its ability to recover when the initial evidence is incomplete or off-target. AlphaAgent replaces this single-pass design with a bounded retrieval loop inspired by active retrieval strategies, allowing the system to inspect and refine its search intent before committing to an answer. After each attempt, the agent evaluates whether returned snippets, metadata, and paper records collectively align with the user request across four dimensions: material system, property, processing condition, and analytical focus. If any dimension is missing or misaligned, the agent treats the evidence as insufficient and triggers reformulation rather than proceeding with generation. This inspection step is critical in materials science, where a single pass may retrieve passages that mention the correct alloy family yet discuss an unrelated processing route. By validating alignment before generation and capping the total number of attempts, the system avoids both mismatched evidence and uncontrolled search expansion.

\begin{figure*}[t]
\centering
\includegraphics[width=0.98\textwidth]{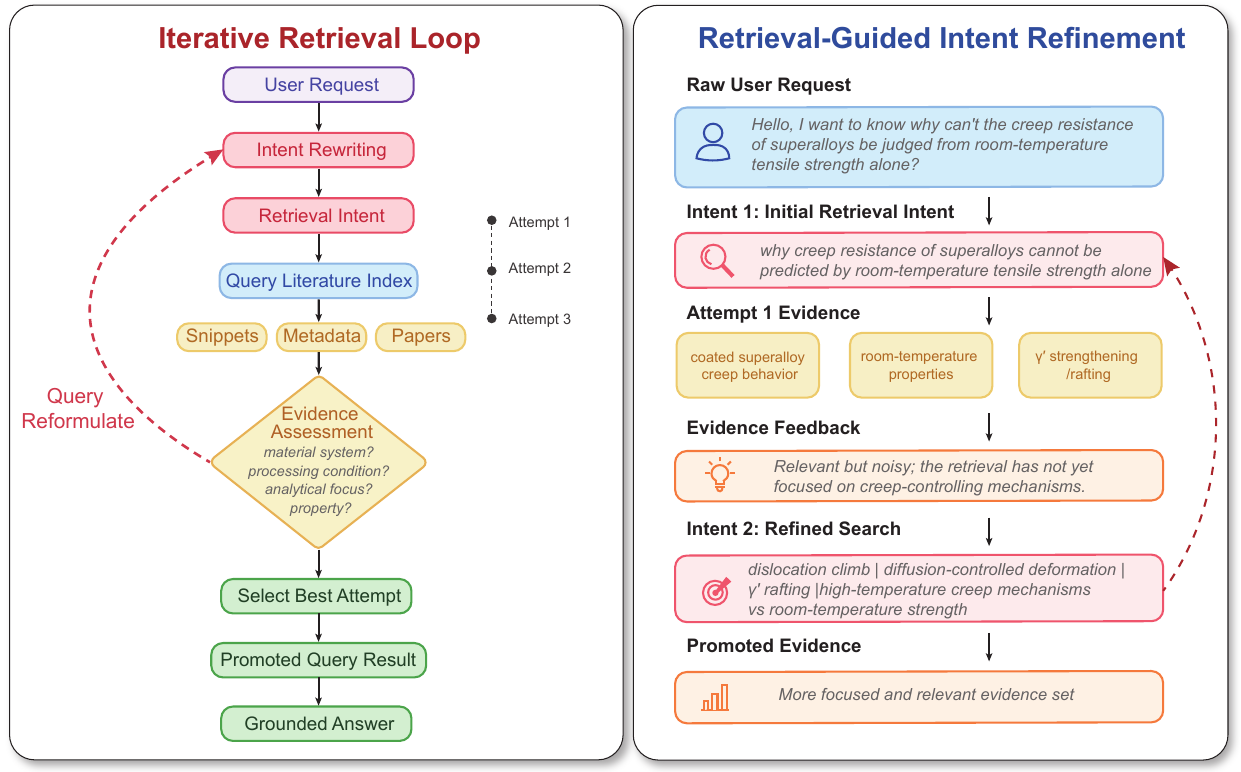}
\caption{Intent-driven iterative retrieval workflow. AlphaAgent rewrites the raw request into a search-oriented retrieval intent, checks whether returned evidence covers the material system, property, processing condition, and analytical focus, and reformulates the query when the evidence is insufficient. The right panel shows an example in which a question about superalloy creep is first re-expressed as a direct search intent, then refined toward mechanism-specific terms after the initial evidence is judged too general. The selected attempt becomes the promoted query result for answer generation.}
\label{fig:iterative_retrieval}
\end{figure*}

If the first attempt falls short, the agent reformulates the retrieval intent and queries the index again, using the observed evidence gap to guide the revision. Reformulation may narrow the material system, add missing processing conditions, or adjust mechanism and property terms so that the next search targets more specific language. The attempt count is bounded to prevent uncontrolled search behavior, and the agent maintains a record of every query and its returned evidence for downstream audit. After exhausting the permitted attempts or finding sufficient evidence, the agent selects the best attempt and promotes it as the query result for answer generation and report preparation. Figure~\ref{fig:iterative_retrieval} shows the retrieval loop, the evidence-assessment criteria, and a concrete example of retrieval-guided intent refinement. This iterative design ensures that the promoted result reflects the most aligned evidence available, rather than the highest-ranked result from an unrefined single pass.

The example in Figure~\ref{fig:iterative_retrieval} illustrates how iterative refinement reshapes the retrieval target without drifting away from the user's original scientific objective. An informal user request asks why the creep resistance of superalloys cannot be judged from room-temperature tensile strength alone, and it is first rewritten into a direct retrieval intent that echoes the question phrasing. The first pass returns general discussions of high-temperature strength but fails to surface the specific deformation mechanisms that differentiate creep from room-temperature behavior. The second attempt adds explicit mechanism language, yet the agent still judges the evidence insufficient because the passages remain too general. The third attempt therefore focuses on concrete microstructural terms, including $\gamma'$ precipitate rafting, dislocation climb, and diffusion-controlled deformation, together with the temperature-versus-strength contrast. This keyword-dense intent finally yields a targeted evidence set that directly supports mechanism-oriented answer generation.

\subsection{Evidence Use in Answer Generation}

Answer generation draws exclusively on the promoted query result, which contains snippets, paper records, document references, and citation metadata from the selected retrieval attempt. The retrieval skill does not read arbitrary PDFs during this stage, because detailed document analysis is reserved for the report-generation skill described in the next section. This strict boundary keeps retrieval-based answer generation lightweight and makes answer sources explicit, preventing the model from silently importing background knowledge that was never retrieved. It also preserves a clear audit trail, since every claim in the answer can be traced back to a specific query result and its constituent snippets. This separation is necessary because it forces the model to ground its reasoning in the returned evidence rather than in parametric memory that often contains outdated or unsupported claims. Without this boundary, retrieval-based answers risk conflating retrieved passages with generative hallucination, which compromises the integrity of the final answer.

The final answer addresses both the original user question and the promoted query result, weaving the retrieved evidence into a coherent response that remains faithful to the source material. When the returned evidence is narrow or incomplete, the answer reflects that boundary rather than filling gaps with unsupported background knowledge. This discipline matters particularly for materials questions, where processing routes, microstructural states, or testing conditions determine whether a reported mechanism applies. By keeping retrieval intent explicit, iteration bounded, and evidence use disciplined, the retrieval skill produces grounded answers that remain strictly within the evidence boundary. The system also surfaces the limitations of the retrieved evidence explicitly, so that users can judge for themselves whether the answer meets their analytical needs. When the promoted evidence demands deeper document-level analysis, the report-generation skill takes over, extending the validated paper set into structured per-paper and cross-paper reports.

\section{Structured Deep Reading and Cross-Literature Synthesis}

\subsection{Query-Linked Paper Selection}

The report-generation skill requires a valid retrieval result. It does not perform cold-start retrieval or summarize snippets directly; instead, it uses paper references and document identifiers from the promoted query result to identify papers for analysis. This keeps the report stage tied to the same evidence-selection process used for the original question.

For each selected paper, the runtime resolves the associated PDF and prepares a paper-specific workspace containing parsed document material, extracted figures and tables when available, and a brief representation for report authoring. Papers that cannot be resolved or parsed are skipped without invalidating reports for other papers.

\subsection{Single-Paper Report Authoring}

The single-paper stage produces a semantic draft organized around identity, scientific content, evidence boundary, and reader plans. In concrete report fields, these categories cover the title, summary, lead, key results, analysis blocks, innovations, evidence-scope notes, conclusion, and plans for figures, tables, highlights, and knowledge structure. These fields ensure coverage of both scientific content and reader-facing organization.

The agent writes the draft from the prepared paper evidence package, identifying the central research question, summarizing main experimental or computational evidence, explaining mechanisms proposed by the authors, and stating the material systems, processing conditions, and evidence ranges supporting the interpretation. The runtime then runs a contract check on the draft and renders the corresponding HTML report only after the draft passes. The check verifies required semantic fields, non-empty reader-facing text, evidence-boundary notes, analysis and key-result blocks, figure/table and reader-plan consistency with the prepared workspace, and the absence of duplicated template prose or leaked raw markup. Invalid drafts enter targeted repair rather than silent rendering: the repair feedback identifies the invalid fields, and the authoring step rewrites those fields while preserving already valid content.

\subsection{Cross-Paper Summary}

When enough valid single-paper reports succeed, AlphaAgent creates a query-level summary. The summary stage reads the original retrieval answer, batch report record, and valid rendered HTML reports without rereading PDFs or introducing new external sources. Its output contains a cross-paper interpretation, structured topic map, theme labels, and a reading path linking back to individual reports.

\begin{figure*}[t]
\centering
\includegraphics[width=0.86\textwidth]{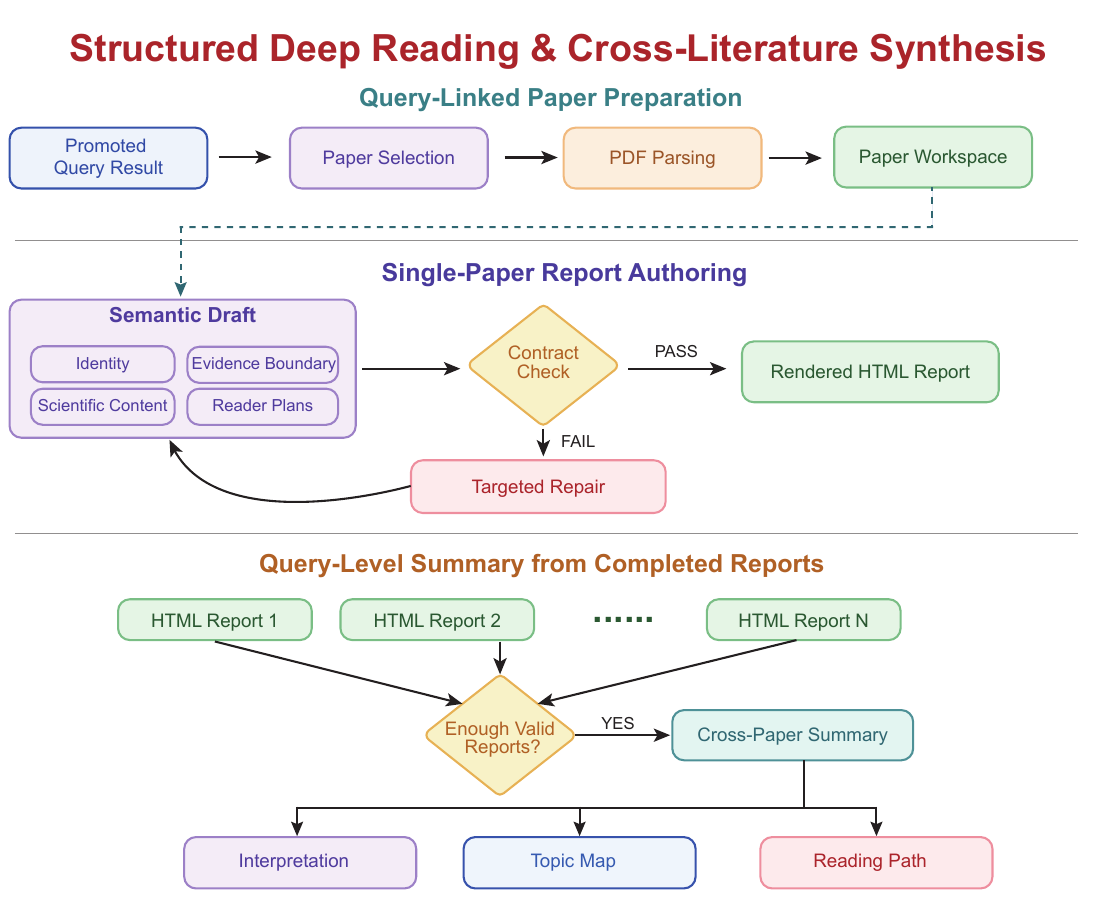}
\caption{Structured deep reading and cross-literature synthesis workflow. The report-generation skill reuses the promoted retrieval result for query-linked paper preparation, parses PDFs into paper workspaces, writes semantic drafts, and runs a contract check before rendering HTML reports. When sufficient valid single-paper reports are available, AlphaAgent synthesizes them into a cross-paper summary with an interpretation, topic map, and reading path.}
\label{fig:deep_reading}
\end{figure*}

This stage targets literature-analysis tasks where users need more than a list of relevant papers. Building the summary from completed single-paper reports lets the framework align papers by material system, processing route, characterization method, mechanism, or performance target. The reading path then provides a practical order for studying the selected literature. Figure~\ref{fig:deep_reading} illustrates the document-level workflow from query-linked paper selection to validated single-paper reports and cross-paper synthesis.

\section{Experimental Evaluation}

\subsection{Evaluation Protocol}

We conducted a blind evaluation in which domain experts rated system outputs without knowing which system produced each answer. The deep analytical question set comprises 20 open-ended analytical questions requiring multi-step mechanistic reasoning, trade-off analysis, and synthesis across multiple studies. These questions probe microstructure--property relationships, processing--structure evolution, and performance limitations in copper alloys, aluminum alloys, titanium alloys, nickel-base superalloys, high-entropy alloys, and additively manufactured metals. A representative deep analytical question asks: In Al$_x$CoCrFeNi high-entropy alloys, increasing Al content often drives FCC to BCC/B2 or multi-phase microstructures. Why might tensile ductility and damage tolerance decrease while hardness increases? The general conceptual question set comprises 20 conceptual questions requiring accurate factual recall, clear mechanistic explanation, and appropriate boundary awareness. A representative general conceptual question asks: Why does ``multi-principal-element'' in high-entropy alloys not necessarily lead to high performance? Under what circumstances can it introduce new microstructural and processing challenges? This evaluation measures answer quality for generated responses, not the rendered single-paper reports or cross-paper report artifacts.

Four systems were evaluated: AlphaAgent, the proposed skill-driven framework with retrieval-intent rewriting, iterative retrieval, promoted-attempt selection, and evidence-grounded answer generation; Baseline RAG, a conventional single-pass pipeline without skill decomposition or query refinement; GPT-5.5, a general-purpose large language model without access to the curated index; and Kimi-K2.6, a general-purpose large language model without access to the curated index. AlphaAgent and Baseline RAG used the same underlying LLM, document index, and retrieval scale, which isolates the effect of skill routing and retrieval-intent refinement rather than differences in the base model or corpus. The Baseline RAG system submits one retrieval query per question, retrieves the same number of evidence items, and provides snippets and metadata to the same LLM. It does not rewrite the user request into a materials-specific intent, inspect whether the returned evidence covers the material system and property trade-off, reformulate the query, or select among multiple attempts. It also does not invoke the report-generation skill.

For each question, reviewers with doctoral-level training in materials science received four anonymized answers labeled A--D in randomized order. Each answer was scored on five criteria using a 1--5 Likert scale: scientific accuracy (correctness of facts, mechanisms, terminology, and causal relationships); question coverage (completeness including sub-questions and implicit constraints); mechanistic depth and trade-off reasoning (explanation of mechanisms and trade-offs rather than fact listing); credibility and boundary awareness (avoidance of overclaiming, unsupported speculation, and excessive certainty, and distinction between established facts and conditional interpretations); and readability and presentation quality (clarity, organization, and fluency of exposition). The overall score for each answer is the arithmetic mean of the five criterion scores, providing a single composite measure of answer quality for subsequent comparison.

\subsection{Overall Performance}

\begin{figure}[t]
\centering
\includegraphics[width=\columnwidth]{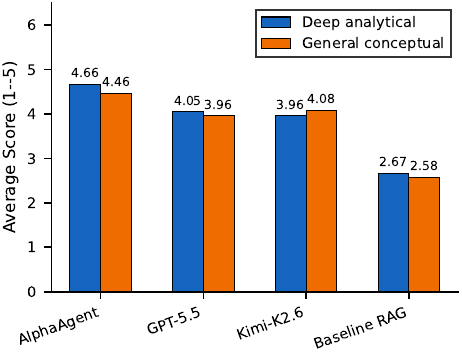}
\caption{Overall average scores across the two task types. AlphaAgent leads on both deep analytical questions and general conceptual questions, with the largest margin observed on deep analytical questions.}
\label{fig:overall}
\end{figure}

Table~\ref{tab:overall_scores} reports average scores for each system and task type. AlphaAgent achieves the highest overall scores: 4.66 on deep analytical questions and 4.46 on general conceptual questions. Baseline RAG scores lowest (2.67 and 2.58), indicating that retrieval over the same corpus benefits from intent refinement and evidence selection. Figure~\ref{fig:overall} visualizes these results. The performance gap between AlphaAgent and the strongest baseline is larger for deep analytical questions ($+0.61$ over GPT-5.5) than for general conceptual questions ($+0.38$ over Kimi-K2.6). This pattern points to the greater importance of retrieval-intent preservation and evidence selection when the task requires mechanism-oriented interpretation rather than short factual explanation. Structured retrieval workflows appear to provide the greatest benefit when questions demand synthesis across multiple experimental conditions and material systems.

\begin{table}[t]
\centering
\caption{Overall Average Scores (1--5 Scale)}
\label{tab:overall_scores}
\begin{tabular}{@{}lcc@{}}
\toprule
System & \shortstack{Deep Analytical\\Questions} & \shortstack{General Conceptual\\Questions} \\
\midrule
AlphaAgent & \textbf{4.66} & \textbf{4.46} \\
GPT-5.5 & 4.05 & 3.96 \\
Kimi-K2.6 & 3.96 & 4.08 \\
Baseline RAG & 2.67 & 2.58 \\
\bottomrule
\end{tabular}
\end{table}

\subsection{Dimensional Analysis}

\begin{figure}[t]
\centering
\includegraphics[width=\columnwidth]{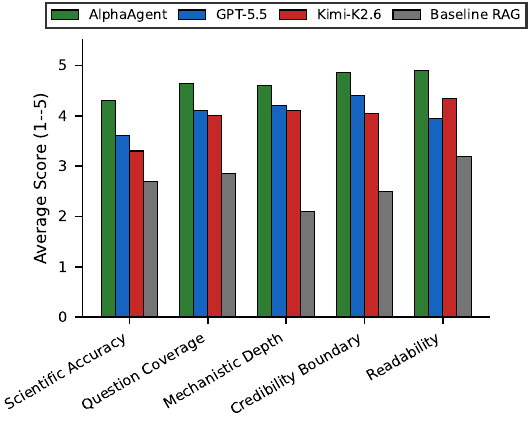}
\caption{Dimension-wise comparison on the deep analytical question set. AlphaAgent shows particularly strong performance on mechanistic depth and credibility boundary awareness.}
\label{fig:deep_dim}
\end{figure}

\begin{figure}[t]
\centering
\includegraphics[width=\columnwidth]{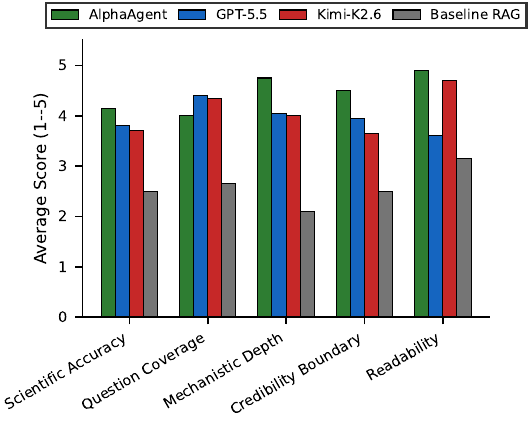}
\caption{Dimension-wise comparison on the general conceptual question set. Performance is more tightly clustered, with AlphaAgent maintaining a consistent advantage in credibility boundary awareness and readability.}
\label{fig:general_dim}
\end{figure}

Figures~\ref{fig:deep_dim} and~\ref{fig:general_dim} break down performance by the five scoring criteria for deep analytical and general conceptual questions, respectively. On the deep analytical question set, AlphaAgent performs most strongly in mechanistic depth and trade-off reasoning (4.60) and credibility and boundary awareness (4.85). These dimensions reflect the retrieval skill's preservation of material systems, phase names, processing conditions, and property trade-offs during query reformulation and answer generation. Baseline RAG scores lowest across all dimensions, with the lowest score in mechanistic depth (2.10), which suggests that retrieved snippets alone provide limited structure for complex analytical answers. On the general conceptual question set, the performance spread is narrower. AlphaAgent leads in scientific accuracy (4.15), mechanistic depth (4.75), credibility boundary awareness (4.50), and readability (4.90), while GPT-5.5 achieves the highest question coverage (4.40). This pattern indicates that retrieval-intent refinement and evidence selection benefit both task types, with the largest gains appearing in tasks that require mechanism-oriented interpretation.

\subsection{Task-Type Differentiation}

The evaluation reveals clear task-type differentiation. AlphaAgent's advantage over the best baseline is larger on deep analytical questions ($+0.61$ over GPT-5.5) than on general conceptual questions ($+0.38$ over Kimi-K2.6), as expected: deep analytical questions demand stronger material-specific retrieval intent, evidence selection, and mechanism-oriented answer organization, while general conceptual questions rely more on concise retrieval and generation. Baseline RAG scores lowest on both task types (2.67 on deep analytical questions, 2.58 on general conceptual questions) even though it uses the same document index and retrieval scale as AlphaAgent. This result suggests that evidence organization and downstream task procedures affect answer quality. Qualitative inspection of the baseline outputs suggests that the gap arises from two patterns. Some answers read as reference-like summaries of locally retrieved snippets instead of mechanism-centered responses. Others draw on neighboring but mismatched examples, such as unrelated alloy families or processing cases, because the single retrieval pass returns evidence without verifying alignment with the requested material system and property trade-off.

In several materials-design questions, the baseline identifies broad experimental tools or literature examples but conflates variables that the question asks to distinguish, such as dislocation strengthening, precipitation strengthening, and residual-solute scattering. These observations are consistent with retrieval drift and snippet over-conditioning. Overall, AlphaAgent provides the greatest value when user requests require structured synthesis across multiple studies rather than simple location of relevant passages. In deep analytical questions, the system benefits from preserving the link between the original question, rewritten retrieval intent, promoted query result, and final explanation. For short general conceptual questions, the performance advantage narrows because these tasks place fewer demands on multi-factor mechanistic synthesis, an area where AlphaAgent's iterative retrieval and evidence control add the most value.

\section{Conclusion}

We introduced AlphaAgent, a skill-driven framework that decouples retrieval-based question answering from document-level report generation through explicit skill contracts, preserving material-specific search intent and enabling structured single-paper and cross-paper analysis. A blind evaluation on forty materials-science questions showed that AlphaAgent substantially outperformed a matched baseline on deep analytical tasks requiring mechanistic explanation, trade-off reasoning, and awareness of credibility boundaries. It also achieved strong performance on general conceptual questions. These findings indicate that explicit task separation, refined retrieval intent, and disciplined evidence use improve the reliability of large-language-model-based literature analysis for materials research. The same contract-based design principles are likely to benefit other scientific domains in which evidence precision and interpretive chain integrity are critical.

\section*{Acknowledgment}

This work was supported by the Tianmushan Laboratory Research Project (TK2024D006, TK2023C021), and the Open Project Program of State Key Laboratory of Artificial Intelligence for Materials Science (Grant Nos. 2025B03 and 2024B04). The authors thank the National Key Laboratory of Artificial Intelligence for Material Science for providing computing resources and platform support.

\bibliographystyle{ieeetr}
\bibliography{ref}

\end{document}